\documentclass[10pt,a4paper]{article}
\usepackage[left=1.2in,right=1.2in,top=.9in,bottom=1.2in]{geometry}
\usepackage{expex}
\usepackage[hidelinks]{hyperref}
\usepackage{graphicx}
\usepackage[final]{microtype}
\usepackage{array}
\usepackage{authblk}
\usepackage{amsmath}
\usepackage{fancybox}
\usepackage{caption}
\usepackage{subcaption}
\usepackage{forest,circledsteps}

\usepackage[
  style=apa,
  natbib=true,
  annotation=false,
]{biblatex}

\bibliography{mpt-multi}

\usepackage{float} 

\title{LLM surprisal is necessary but not sufficient to capture English garden-path effects: Evidence from joint latent modeling of reading paradigms\thanks{This work was funded by the Deutsche Forschungsgemeinschaft (DFG, German Research Foundation) --- Project number 428960187 --- PI: Dario Paape.}}

\author[1,*]{\mbox{Dario Paape}}
\author[2]{\mbox{Tal Linzen}}
\author[1]{\mbox{Shravan Vasishth}}
\affil[1]{Department of Linguistics, University of Potsdam}
\affil[2]{Center for Data Science/Department of Linguistics, New York University}
\affil[*]{Please send correspondence concerning this article to: Dario Paape, Department Linguistik, Universität Potsdam, Karl-Liebknecht-Straße 24--25, D-14476, Potsdam, Germany. Email: \href{mailto:paape@uni-potsdam.de}{paape@uni-potsdam.de}.}

\begin{document}

\maketitle

\begin{abstract}
Temporarily ambiguous garden-path sentences (\textit{While the team trained the striker wondered \ldots}) are known to cause processing difficulty, which can manifest itself in a variety of reading behaviors (in-situ slowdowns, rereading), as well as in miscomprehension or outright rejection of the sentence as ungrammatical. Which types of reading behavior are observed critically depends on the experimental method used to collect the data, which makes comparing results between reading paradigms difficult. To address this problem, we present a latent-process multinomial processing tree (MPT) model of human reading and comprehension/judgment behavior in garden-path sentences that we fit to combined data from four different reading paradigms (eye tracking, uni- and bidirectional self-paced reading, Maze). The model distinguishes between the probability of adopting an incorrect initial analysis, the cost of encountering an incompatible continuation, and the cost of syntactic reanalysis. By taking into account trials with inattentive reading, more realistic estimates of the cost parameters are obtained. Cross-validation reveals that the MPT model has a better predictive fit to human reading patterns and end-of-trial task data than a model based solely on LLM-derived surprisal values. We also test several models that assume an influence of surprisal within the MPT architecture, and find that adding surprisal as an additional predictor or reading time and/or garden-path cost further improves predictive fit.

\end{abstract}

\section*{Background}

One of the most enduring questions in psycholinguistics concerns the precise mechanisms involved in human sentence processing, especially the real-time selection and maintenance of syntactic analyses. In this context, particular attention has been paid to so-called garden-path sentences. Garden-pathing occurs when the structure of a given sentence is disambiguated towards an initially dispreferred analysis, and can occur in a variety of syntactic contexts. One type of garden-path context that we will study is the NP/Z (zero complement) garden path, which is exemplified in (\ref{x1}).

\ex\label{x1}While the team trained(,) the striker wondered whether the damage would heal.
\xe

NP/Z garden paths are usually relatively severe and tend to cause noticeable processing difficulty \citep{sturt1999structural}. In example (\ref{x1}), when the comma is absent, most readers first tend to combine \textit{the striker} with \textit{trained}. However, when \textit{wondered} is encountered, this analysis is rendered impossible, as the main clause needs a subject, and \textit{trained} must be reanalyzed as intransitive (as in the comma version). Since the 1970s, a range of different processing theories have been proposed to explain the processing difficulty that readers experience in the no-comma version of such sentences. These theories differ along many dimensions: whether the parser is assumed to be serial or parallel, whether its initial structural choice is probabilistic or deterministic, and which factors are assumed to influence garden-path strength (e.g., \citealp{frazier1982making,waltz1985massively,milne1982predicting,clifton1989ambiguity}). Since garden-path sentences tend to produce rich data in the sense that they show a large variety of different reading behaviors (e.g., \citealp{von2011scanpath}), they are also a fruitful target for complex analyses and computational modeling of reading patterns. 

Crucially for the purposes of the present paper, the recent rise of large language models (LLMs) has reignited a debate about whether human reading behavior in garden-path sentences is categorically different from more ``normal'' sentences, in the sense that specialized cognitive processes such as a reanalysis module need to be stipulated (e.g., \citealp{van2021single}). An alternative possibility is that the difficulty of garden-path sentences reduces to the disambiguating word (\textit{wondered} in (\ref{x1})) being highly unexpected given the previous linguistic context (e.g., \citealp{levy2008expectation}). 

Assuming that, in any given sentence, a word's processing difficulty is proportional to its unexpectedness, and given that word probabilities derived from LLMs can serve as a useful proxy for a word's probability in context (e.g., \citealp{wilcox2023testing}), it should be possible to bring garden-path sentences ``into the fold'' without having to assume additional cognitive machinery: if an LLM assigns a low probability to \textit{wondered} in (\ref{x1}), this should be sufficient to explain why readers have trouble integrating it. However, a number of recent studies have cast doubt on this proposal, finding that the processing disruption in garden-path sentences goes beyond what is explainable by LLM-derived surprisal \citep{van2021single,huang2024large,timkey_huang_oh_prasad_arehalli_linzen_dillon_2025,pvc}. Nevertheless, a surprisal-based account of garden-pathing remains an attractive possibility due to its theoretical parsimony, and because surprisal has been shown to predict human processing difficulty in a variety of other linguistic constructions \citep{staub2025predictability}. 

Ultimately, decisions about competing theories must be driven by data --- the more, the better. Informative data with regard to garden-pathing include human reading times when first encountering the disambiguating word, but also measures of rereading, which signal ongoing processing difficulty, as well as data from explicit end-of-trial tasks such as comprehension questions and acceptability judgments. In the current work, we fit a computational model that assumes dedicated recovery processes and compare its predictive performance to that of pure LLM-derived surprisal. Based on a large set of existing reading data from different reading paradigms, we conclude that dedicated processes \textit{are} needed to adequately account for the comprehension of garden-path sentences.

\subsection*{The case for modeling shared latent processes across reading paradigms}

The most well-known paradigm used to study human reading behavior, including the reading of garden-path sentences, is eye tracking during reading (e.g., \citealp{rayner2006eye}). Another widely-used method is self-paced reading (SPR), in which participants reveal a masked stimulus text incrementally via key presses on a computer keyboard, and the time between key presses is recorded \citep{just1982paradigms}. In classic self-paced reading, rereading is not possible, as words are re-masked once the participant has moved past them. Recently, \citet{paape2022reanalysis} have developed a variant of the paradigm called bidirectional self-paced reading (BSPR) that allows participants to move back and forth through the sentence, thus making it possible to reread earlier words. Another relatively new method is the Maze paradigm \citep{forster2009maze,boyce2023maze}, which requires the participant to incrementally ``build up'' the sentence: at each word, the participant has to choose (via key presses) between between a grammatical continuation and an ungrammatical alternative or non-word. Maze does not allow rereading, and the trial typically ends immediately if an incorrect choice is made. An advantage of Maze is that the nature of the task forces incremental processing to be highly localized, largely eliminating processing spillover to subsequent words. Maze studies are also argued to eliminate the need for end-of-trial comprehension questions, as choosing the correct continuation implies that the sentence was understood correctly.

A crucial assumption made in psycholinguistic research that is rarely made explicit is that all reading paradigms ultimately measure the output of a shared set of cognitive processes that, taken together, constitute sentence processing. That is, modulo differences related to motor and eye movement control, parafoveal preview of upcoming words, and the possibility of rereading previous words, the measures generated by the different methods should be influenced by linguistic manipulations in similar ways. Indeed, reading-time measures collected with different experimental paradigms often (but not always) yield converging results for similar experimental manipulations (e.g., \citealp{paape2022conscious,koornneef2006use,frank2013reading,witzel2012comparisons}). 

In work on garden-path processing, there is converging empirical evidence across different reading paradigms that the ambiguous no-comma version of (\ref{x1}) is more difficult to read at the point of disambiguation than the comma version, as indexed by longer reading times and/or increased rereading in paradigms where rereading is possible. The no-comma version also tends to cause lingering misinterpretations even after disambiguation, such that participants presented with the question \textit{Did the team train the striker?} often answer \textit{yes}. Finally, the no-comma version is more likely to be rejected as ungrammatical, suggesting that participants sometimes give up on trying to parse the sentence (see \citealp{paape2022reanalysis} for a review). 

Despite these converging findings, their interpretation in terms of the cognitive processes involved, especially across methods and measures, has remained contentious. For instance, there is disagreement as to whether syntactic reanalysis is separate, identifiable processing stage handled by a specialized module or simply handled by ``normal'' parsing mechanisms (e.g., \citealp{McRae1998,van2021single}). Assuming that a reanalysis module exists, there are also diverging proposals with regard to its operating principles; for instance \citet{frazier1982making} have suggested that reanalysis involves targeted rereading of earlier sentence portions, while \citet{Lewis98} assumes that reanalysis can be carried out ``covertly'' by slowing down at the disambiguating region and backtracking in memory.

There has, to our knowledge, been no attempt to build a computational model that simultaneously captures reading behavior across all of the aforementioned experimental paradigms. For eye movement data, there is a wealth of highly complex computational models that aim to capture reading behavior with high granularity (e.g., \citealp{engbert2005swift,reichle2003ez,snell2018ob1}). \citet{nicenboim2025cofi} has recently proposed a computational model for self-paced reading that borrows many ideas from eye-movement models. The disadvantage of building detailed models of reading behavior in one paradigm, however, is that the models cannot be readily extended to other reading paradigms. The model we present below aims to fill this gap by extending a recent approach by \citet{paape2022estimating,pvc,paape2026deconstructing}, which uses multinomial processing trees (MPTs; see \citealp{erdfelder2009multinomial,riefer1988multinomial} for reviews) as the basic architecture.

Outside of computational modeling work, the standard approach in the field is to take data from a single paradigm and compare the mean reading times of target and control sentences across many participants. Depending on the experimental setup, researchers also analyze regression probabilities and/or the proportion of positive acceptability judgments or correct question responses in a separate statistical model. In our running example (\ref{x1}), researchers would typically assume the cognitive cost of garden-pathing to be indexed by differences between the comma and no-comma versions in mean reading times, regression probabilities and/or the proportion of trials with incorrectly answered comprehension questions or outright rejections of the sentence as ungrammatical. The difference in mean reading times, in turn, is often taken to index the time taken for syntactic reanalysis, that is, the cost of ``undoing'' the initial incorrect syntactic attachment and building the correct structure, or as a reranking of syntactic analyses computed in parallel (e.g., \citealp{meng2000ungrammaticality}). Increases in rejection and/or misinterpretation probability are usually interpreted as reanalysis failures (e.g., \citealp{warner1987context,christianson2001thematic}).

There are two crucial drawbacks to the standard analysis approach of comparing means for each measure of processing difficulty in isolation, and within a single reading paradigm. First, analyzing reading times and rejection/misinterpretation probabilities separately ignores the fact that both kinds of measures plausibly arise from the same set of cognitive processes \citep{huang2021causes}. Computational modeling in cognitive psychology typically takes into account the connection between responses and their associated reaction times, and the reaction time is often taken to represent the amount of time taken to accumulate enough information to give a response (e.g., \citealp{ratcliff2006modeling}). In psycholinguistics, outside of computational modeling approaches (e.g., \citealp{nicenboim2018models,o2026drift,HAMMERLY201970}), a distinction is often made between ``online'' measures such as reading times and ``offline'' measures such as end-of-trial judgments, and mismatches between the two in terms of sensitivity to experimental manipulations are often given a theoretical explanation (e.g., \citealp{pynte1997syntactic,bader2023processing}). In the computational model we present below, we will follow the computational approaches in cognitive psychology and assume a direct connection between reading times and end-of-trial responses.

Second, by only comparing average reading times between conditions, researchers assume that there is one common distribution that generates all observations \citep{ratcliff2002estimating}. This assumption glosses over the possibility that combinations of different processes may occur or not occur in different experimental trials. A straightforward solution to this problem is to assume a mixture distribution of reading times (e.g., \citealp{vasishth2017feature,vasishth2017modelling,nicenboim2018models}).

For instance, participants may not always be fully attentive during reading, perhaps due to insufficient motivation \citep{christianson2022if}, and may only quickly ``skim'' the sentence (e.g., \citealp{mills2017cognitive}), resorting to guessing when asked a comprehension question. In a garden-path sentence such as (\ref{x1}), assuming that the participant fully engages with the reading task, they may carry out costly reanalysis procedures that can involve backtracking through the sentence, as indexed by a marked slowdown at the disambiguating region (``covert'' reanalysis) or by overt rereading \citep{Lewis98}. Alternatively, a reader may notice the unexpectedness of \textit{wondered} and respond with a slight reading slowdown, but postpone reanalysis and keep going forward \citep{keshev2020effects}, or not attempt reanalysis at all. Finally, even though there is a strong tendency to initially misanalyze garden-path sentences, there may be trials in which the correct analysis computed from the get-go even without the comma (e.g., \citealp{jurafsky1996probabilistic,van2000unrestricted}). A plausible computational model of garden-pathing --- as well as any general model of reading that includes linguistic processing beyond word recognition --- should incorporate all of these possibilities. 

With regard to the surprisal-versus-reanalysis debate, earlier work used differences between average reading times across conditions to argue against a pure surprisal account \citep{van2021single,huang2024large}, but more recent work has distinguished between trials with rereading versus trials without it \citep{timkey_huang_oh_prasad_arehalli_linzen_dillon_2025}, or used joint analysis of dependent variables and mixture modeling to argue for a multi-stage cascade of latent processes \citep{pvc,paape2026deconstructing}. A crucial advantage of the latent-process approach is that it is paradigm-independent. Unlike eye-tracking data, which lends itself well to a separation into ``early'' versus ``late'' measures of processing in terms of first-pass reading versus rereading (but see \citealp{vasishth2013eye} for criticism of this mapping), data from unidirectional self-paced reading do not naturally allow for such a separation. Yet, it is plausible to assume that the same processes that drive rereading of garden-path sentences in eye tracking are also behind the longer reading times that are usually observed in self-paced reading. Below, we follow the approach of \citet{pvc,paape2026deconstructing} to model reading data across paradigms, with the aim of uncovering the shared latent processes generating the data.

\section*{The multinomial processing tree model of garden-pathing}

The multinomial processing tree (MPT) model was initially proposed by \citet{paape2022estimating} and extended by \citet{pvc,paape2026deconstructing}. The MPT model treats garden-pathing as nondeterministic, which implies that reading times should follow a mixture distribution. The model assumes five mixture components for reading times. They are distinguished by whether the reader is currently attentive or inattentive, whether garden-pathing occurs or does not occur, and whether reanalysis is carried out ``covertly'' in-situ or via rereading. The shift parameter encodes the assumption that part of the total measured reaction time is non-decision time, that is, the time required to encode the stimulus, execute motor actions, and so forth (e.g., \citealp{rouder2005unshifted,verdonck2016factoring}). 

\begin{itemize}\small
  \item[] \Circled{1}~~If not reading attentively:\vspace{-.5ex}
  \item[] \hspace{.5cm}$(\text{RT}-\text{shift}) \sim LogNormal(\mu,~\sigma_1)$\vspace{.3ex}
  \item[] \Circled{2}~~If reading attentively but not garden-pathed:\vspace{-.5ex}
  \item[] \hspace{.5cm}$(\text{RT}-\text{shift}) \sim LogNormal(\mu+att\_cost,~\sigma_2)$\vspace{.3ex}
       \item[] \Circled{3}~~If garden-pathed, no in-situ (covert) reanalysis:\vspace{-.5ex}
   \item[] \hspace{.5cm}$(\text{RT}-\text{shift}) \sim LogNormal(\mu+att\_cost+gp\_cost,~\sigma_2)$\vspace{.3ex}
  \item[] \Circled{4}~~If garden-pathed, in-situ (covert) reanalysis:\vspace{-.5ex}
   \item[] \hspace{.5cm}$(\text{RT}-\text{shift}) \sim LogNormal(\mu+att\_cost+gp\_cost+reanalysis\_cost,~\sigma_2)$\vspace{.3ex}
       \item[] \Circled{5}~~If garden-pathed, regression triggered (only in eye tracking, BSPR):\vspace{-.5ex}
   \item[] \hspace{.5cm}$(\text{RT}-\text{shift}) \sim LogNormal(\mu+att\_cost+gp\_cost+regression\_cost,~\sigma_2)$
\end{itemize}

The MPT model also assumes the following latent processes with their associated probabilities:

\begin{enumerate}
\item Attention ($p\_attentive$): The probability that the participant is reading attentively. Varies between participants.
\item Garden-pathing ($p\_gp$): The probability that the participant is garden-pathed, that is, the disambiguating word cannot be integrated into the current analysis. Varies between conditions (garden-path vs.~control, garden-path type), participants, and sentences.
\item Overt reanalysis ($p\_overt$): The probability that the participant engages in overt reanalysis (rereading). If there is no overt reanalysis, it is assumed that covert, in-situ reanalysis is carried out instead, resulting in a local slowdown. Varies between participants.
\item Postpone reanalysis ($p\_postpone$): If reanalysis is covert, it can be postponed from the critical disambiguating region to the following spillover region, so that the slowdown will register there. Varies between participants.
\item Reanalysis success ($p\_success\_o$/$p\_success\_c$): The probability that overt/covert reanalysis is successful. Varies between garden-path types (see below).
\item Pragmatic inference ($p\_infer$): The probability that the participant engages in pragmatic inference when answering a comprehension question. For instance, when being asked \textit{Did the team train the striker?} about (\ref{x1}), the participant may plausibly infer that the team did train the striker, even though this was not explicitly stated (e.g., \citealp{christianson2001thematic,ferreira2001misinterpretations}). Varies between participants and sentences.
\item Baseline regression probability ($p\_base\_regress$): Participants may make regressions even in the absence of garden-pathing in order to correct for motor errors, to confirm their interpretation, or to ``wrap up'' unfinished processing \citep{christianson2017reread,reichle2009using,andrews2021wrapping}. Varies between participants.
\end{enumerate}

The multinomial component of the model is informed by the presence or absence of rereading in the trial after having read the critical region, as well as by the acceptability judgment (\textit{accept}/\textit{reject}) or comprehension question response (\textit{Did the team train the striker?}; \textit{yes/no}). Judgments and question responses are registered at the end of the trial and are assumed to index the final outcome of processing. Trials are grouped into four categories according to the possible combinations of these two binary variables (\textit{accept/no, no regression; accept/no, regression; reject/yes, no regression; reject/yes, regression}). Correctly accepting the sentence as grammatical or correctly answering a comprehension question about the initial misparse with \textit{no}, depending on the task used in the experiment, are assumed to reflect successful parsing. By contrast, rejecting the sentence as ungrammatical or incorrectly answering \textit{yes} are assumed to reflect unsuccessful parsing. 

The four outcomes are modeled with a multinomial distribution whose probabilities are determined by the structure of the processing tree. The outcome probabilities are connected to the mixture distribution of reading times via the mixing proportions, which allows us to estimate process probabilities and process costs simultaneously (see also \citealp{heck2018generalized}). The model is hierarchical with crossed random effects, that is, all costs and probabilities can vary between participants and sentences where appropriate.

The model of \citet{paape2026deconstructing} contains two simplifications compared to earlier versions. First, \textit{accept/reject} or \textit{yes/no} responses are generated with equal probability under guessing, that is, there is no response bias parameter. Second, the model does not assume a ``triage'' process that leads to quick rejections instead of reanalysis (e.g., \citealp{fodor2000garden,paape2022estimating}). We make this assumption because \citet{pvc} found that a model with triage did not show improved fit over a model without triage.  Figure \ref{tree} shows the extended version of the MPT model that we will be referring to throughout.

A novel addition to the MPT model of \citet{paape2026deconstructing} compared to the model of \citet{pvc} is the $p\_infer$ parameter that captures pragmatic inferences. $p\_infer$ accounts for \textit{yes} responses to comprehension questions that are not informative about the actual parsing outcome. The model assumes that (correct) \textit{accept} judgments and (correct) \textit{no} responses map onto each other modulo the contribution of $p\_infer$ to the question responses. Studies with acceptability judgments and studies with comprehension questions can thus be modeled together, and the amount of pragmatic inferences as a source of misinterpretations can be estimated directly. 

\begin{figure}\centering
\scalebox{0.85}{
\begin{forest}
for tree={grow=south, align=left, font=\scriptsize, l=55pt, s sep=-5pt,
          before typesetting nodes={content=\strut#1}}
[{\textbf{Attentive?} \textcolor{red}{$\leftarrow$\bf~Participant}\\Probability: $p_{attentive}$},
  s sep=-35pt,
  [{\bf \hspace{1cm}Guess\\\hspace{1cm}Probability: $0.5$},
    edge label={node[midway,left,yshift=1ex,xshift=-2ex]{\ovalbox{no}}}
    [{\sc \doublebox{answer ``yes''}}]
    [{\sc \doublebox{answer ``no''}}]
  ]
  [{\bf \hspace{4cm}Garden-pathed? \textcolor{red}{$\leftarrow$ Construction, condition}\\\hspace{4cm}Probability: $p_{gp}$},
    edge label={node[midway,right,yshift=1ex]{\ovalbox{yes} (pay attention cost)}}
    [{\hspace{2cm}\sc \doublebox{infer?}~\textcolor{red}{$\leftarrow$\bf~Participant}},
      edge label={node[midway,left,yshift=2ex,xshift=2ex]{\ovalbox{no}}}
    ]
    [{\bf \hspace{1.5cm}Overt reanalysis? \textcolor{red}{$\leftarrow$ Construction, paradigm}\\\hspace{1.5cm}Probability: $p_{overt}$},
      edge label={node[midway,right,yshift=1ex]{\ovalbox{yes} (pay GP cost)}}
      [{\bf \hspace{2.5cm}Postpone? \textcolor{red}{$\leftarrow$ Construction, paradigm~~~~~~~~~~~~~~~}\\\hspace{2.5cm}Probability: $p_{postpone}$},
        edge label={node[midway,left,yshift=1ex]{\ovalbox{no}}}
        [{\textbf{Success?}\\Probability: $p_{success\_c}$},
          edge label={node[midway,left,yshift=1ex]{\ovalbox{no} (pay cost at crit)}}
          [{\sc \doublebox{infer?}},
            edge label={node[midway,left,yshift=1ex]{\ovalbox{yes}}}
          ]
          [{\sc \doublebox{answer ``yes''}},
            edge label={node[midway,right,yshift=1ex]{\ovalbox{no}}}
          ]
        ]
        [{\textbf{Success?}\\Probability: $p_{success\_c}$},
          edge label={node[midway,right,yshift=.5ex,xshift=3ex]{\ovalbox{yes} (pay cost at crit+1)}}
          [{\sc \doublebox{infer?}},
            edge label={node[midway,left,yshift=1ex]{\ovalbox{yes}}}
          ]
          [{\sc \doublebox{answer ``yes''}},
            edge label={node[midway,right,yshift=1ex]{\ovalbox{no}}}
          ]
        ]
      ]
      [{\sc \doublebox{\ldots}},
        edge label={node[midway,right,yshift=1ex]{\ovalbox{yes} (pay regr.~cost)}}
      ]
    ]
  ]
]
\end{forest}}
\caption{The multinomial processing tree model of garden-pathing. The elided portion of the tree under [\ldots] in the overt analysis branch is analogous to the covert reanalysis branch, with $p\_success\_c$ replaced with $p\_success\_o$.\label{tree}}
\end{figure}

Because the latent processes the MPT model assumes should generalize across reading paradigms, it lends itself well to fitting data from eye tracking, uni- and bidirectional self-paced reading, and Maze. Differences between paradigms, such as the possibility or impossibility of regressions, can be accommodated by either constraining the appropriate parameters accordingly or allowing them to vary freely between paradigms. 

For instance, the model is constrained in such a way that SPR and BSPR are assumed have the same amount of non-decision time. Eye tracking is assumed to have a shorter non-decision time than (B)SPR, due to the less deliberate nature of eye movements compared to key presses. Maze, by contrast, is assumed to have a longer non-decision time, as the participant is presented with two continuation options that they have to process and decide between.

Other parameters are intentionally left unconstrained, so that the estimates are mainly driven by the data. For example, the claim that there is less spillover in Maze than in (B)SPR can be evaluated by letting the parameter $p\_postpone$ vary between paradigms and checking whether the slope (difference) estimate is positive or negative. Fitting the MPT model to data from different paradigms thus serves a dual purpose: first, to see if a single model can fit cross-paradigm data well and yield information about the shared latent processes involved, and second, to see if claims made in the literature --- such as Maze having less spillover than SPR \citep{boyce2023maze} --- are substantiated by the relevant parameter estimates.

All cost parameters in the model are constrained to be greater than zero, and $\sigma_1$ is constrained to be smaller than $\sigma_2$, assuming that reading times during inattentive reading are less variable compared to attentive reading \citep{bixler2015automatic,paape2022estimating}. We model reading times at the critical region and the spillover region --- that is, the region directly following the critical region --- to account for postponed reanalysis (see below). As an extension over previous work, we also include reading times from the region directly preceding the critical region; this addition is intended to improve the estimate of baseline regression probability in the absence of garden-pathing, as well as baseline reading time for a region close to the critical region in which no garden-path or reanalysis costs should arise. Reading times in the precritical region are thus assumed to be generated only from components \Circled{1} (inattentive reading) and \Circled{2} (attentive, no garden-pathing) above. Reading times at the precritical, critical and spillover region have the same amount of non-decision time, but non-decision time can vary between participants \citep{nicenboim2018models}.

As regressions are consciously triggered in BSPR  \citep{paape2022reanalysis}, the probability of overt reanalysis and the base regression probability are assumed to be lower in BSPR compared to eye tracking, while the regression cost is assumed to be higher. $p\_overt$ and $p\_base\_regress$ are set to zero for SPR and Maze experiments.

Before proceeding to the model fits, we will now introduce a much simpler competitor model of garden-path effects. The model in question is simply a linear regression based on large language model (LLM) surprisal.

\section*{Surprisal as a more parsimonious competitor model}

Human reading behavior is complex, but a computational model is only as good as its fit to the data, and simpler models should be preferred if they offer a comparable or even better fit. One such model that we will compare against our multi-process model is based on surprisal, that is, the (negative log) probability of a word given the preceding linguistic context \citep{hale2001probabilistic,levy2008expectation}. Surprisal theory assumes that processing difficulty of a word increases linearly with its negative log probability, that is, the more unexpected a word is in a given context, the longer its reading time. Crucially, surprisal theory does not assume qualitatively distinguishable categories of trials with different serial processing paths, such as ``reanalysis'' or ``no reanalysis''. Instead, parallel parsing is assumed --- though the exact amount of parallelism can vary (e.g. \citealp{boston2008surprising}) --- and processing difficulty reflects the redistribution of probability across different analyses. \citet{levy2008expectation} has also proposed that surprisal acts as a ``causal bottleneck''. Under this view, cognitive mechanisms that act on linguistic representations --- such as assigning a particular syntactic analysis --- can only have an effect on ``behavorial observables'' by changing next-word probabilities (see also \citealp{smith2013effect}). 

With the rise of LLMs, it has become relatively trivial to compute surprisal for arbitrary sentences, and LLM-derived surprisal often yields better predictive fit to human reading measures than ``empirical'' surprisal from sentence completion studies (e.g., \citealp{lopes2024language,hofmann2022language}). Surprisal successfully predicts human reading measures across different experimental methods (e.g., \citealp{shain2024large,wilcox2023testing}; see \citealp{staub2025predictability} for a review). This extends to phenomena that were previously thought to be mediated by complex syntactic and/or semanto-pragmatic mechanisms, like the construction of situation models \citep{michaelov2023can} or the interaction between discourse context and structural parsing preferences \citep{pvc}.

Under the ``causal bottleneck'' perspective, the processing difficulty in garden-path sentences like (\ref{x1}), repeated below as (\ref{x1a}), is simply due to the word \textit{wondered} being less predictable when the comma is absent compared to when it is present.

\ex\label{x1a}While the team trained(,) the striker wondered whether the damage would heal.
\xe

Garden-path sentences thus have no special theoretical status under this view, and no specialized processing mechanisms should be needed. Indeed, assuming that ``all types of hierarchical predictive information present in language [...] must inevitably bottom out in predictions about what specific word will occur in a given context'' \citep[p.~313]{smith2013effect}, a linear regression model with word-level surprisal as the sole predictor should theoretically be sufficient to account for reading times irrespective of sentence type. Empirically, however, models that assume a linear relationship between surprisal and reading times underpredict the empirically observed difference between the ambiguous and unambiguous versions of garden-path sentences \citep{van2021single,huang2024large,timkey_huang_oh_prasad_arehalli_linzen_dillon_2025,mainakilaas2026revisiting}. Importantly, \citet{timkey_huang_oh_prasad_arehalli_linzen_dillon_2025}, who used eye tracking, found that surprisal's underprediction was less severe for trials \textit{without} regressions, and argued that trials with and without rereading may index different facets of processing. Data from methods like unidirectional SPR and Maze cannot be neatly partitioned in the same way, and this is one of the issues that the MPT model addresses: if the overt reanalysis route is blocked, covert analysis (local slowdown) is assumed to apply instead, and the total amount of reanalyzed trials should be the same across all paradigms.

\citet{pvc} compared a latent-process MPT model against GPT-2 \citep{radford2019language} surprisal in terms of predictive fit to human garden-path processing data, and found that the MPT model clearly outperformed surprisal. Given the MPT's assumption of trial-by-trial stochastic variability in terms of which cognitive processes occur in response to garden-pathing, a possible conclusion is that it is precisely the LLM's lack of such variability that makes it a non-optimal model of human behavior. On the theoretical side, it is also not clear if the ``causal bottleneck'' view of surprisal has anything to say about attention lapses, postponed reanalysis, or pragmatic inferences. The MPT model explicitly assumes that these factors do affect ``behavorial observables'', but it is unclear how to group them into the categories of mechanisms that surprisal theory assumes to feed into the causal bottleneck of next-word prediction. That being said, it is plausible that, say, occasional attention lapses would affect the surprisal of upcoming words, so that surprisal should not be completely ruled out as a catch-all explanation of processing difficulty.

An unexplored possibility is that surprisal and the latent processes assumed by the MPT model jointly contribute to human garden-path processing. One variant of such a hybrid view would be that surprisal affects process probabilities in the MPT. For instance, surprisal could affect or even fully determine the initial structural commitment of the parser in cases where processing is attentive, that is, act as a slope on the $p\_gp$ parameter. This effect could theoretically be the whole extent of surprisal's contribution; it might contribute nothing to reanalysis probability, reanalysis cost, or possible pragmatic inferences. Another hybrid model is one in which surprisal does \textit{not} affect structural choice but tracks the cost of having made an incorrect commitment, that is, the garden-path \textit{cost} parameter of the MPT model. The latter model would be somewhat more in line with the spirit of the surprisal account, in the sense that it assumes a smooth mapping between surprisal and reading times in the subset of trials with garden-pathing, as opposed to determining which of two categorically distinct processing paths is taken. 

Finally, it could be that including surprisal in the processing model is necessary to account for lower-level processes such as word identification, while higher-level processes such as structural parsing do not interact with surprisal at all. Another possibility is that surprisal does capture \textit{some} higher-level integration processes as well, but has trouble accounting for trials in which these processes fail and laborious reanalysis becomes necessary \citep{timkey_huang_oh_prasad_arehalli_linzen_dillon_2025}. We will test a range of hybrid models below.

\begingroup
\sloppy
\section*{Fitting the models to reading data}
\endgroup

In order to evaluate the models on data from different reading paradigms, we use a convenience sample of open-access reading data from experiments on garden-path processing. We focus on the so-called NP/zero complement (NP/Z) ambiguity illustrated in (\ref{x1}), as well as on main verb/reduced relative (MV/RR) ambiguities, as illustrated in (\ref{x2}), as these are the two most widely studied types of garden paths in the literature.

\pex\label{x2}\a The girl (who was) fed the lamb remained relatively calm.
\a The lawyer/package sent by the governor was neglected by the officials.
\xe

In (\ref{x2}), the first noun (\textit{girl}) is likely to be misinterpreted as the agent of the first verb (\textit{fed}). At the main verb (\textit{remained}), the would-be agent noun has to be reinterpreted as the patient of the verb (\textit{the girl who was fed\ldots}). The ambiguity can be removed by either using an unreduced relative clause as in (\ref{x2}a), or by using an inanimate noun as in (\ref{x2}b). Both types of manipulation are present in our sample. As the magnitude of the garden-path effect may depend on the type of manipulation used, the MPT model contains an additional parameter that allows the garden-path probability $p\_gp$ to vary between disambiguation types. By contrast, in the surprisal model, differences between disambiguation types should already be subsumed by differences in the surprisal of the disambiguating word.

In order to yield good estimates of the latent process probabilities and costs, the MPT model optimally needs grammaticality judgments or comprehension question responses in addition to reading times in each experimental trial. This is because question responses and judgments yield more direct evidence regarding which processing path was taken, whereas reading times are inherently more noisy and thus less informative about which mixture component a given measurement was generated from. This reflects a general property of finite mixture models: when the distributions of different mixture components substantially overlap --- as is the case for reading times generated by adjacent processing paths in our model --- the generating component for a single observation is only weakly identified by the continuous outcome alone \citep{McLachlan2000FiniteMM}. By contrast, some combinations of the binary outcomes, such as a positive grammaticality judgment in the presence of a regression, can only be generated by a limited subset of paths, thereby helping disambiguate which component the observation belongs to.

Relatively few studies in the literature meet these desired criteria. For instance, Maze studies typically do not use comprehension questions, as it is argued that participants have to comprehend the sentence in order to choose the correct continuation \citep{forster2009maze}. For the MPT model, this would imply that all Maze trials are grouped as having been correctly parsed. However, as we do not want to commit to the assumption of perfect accuracy in Maze, we decided to treat Maze accuracy as missing data instead.

We selected the following studies for our sample:\vspace{1ex}

\begin{itemize}
\item \citet{huang2024large}: A large-scale SPR study on NP/Z and MV/RR garden paths (n = 2000) known as the syntactic ambiguity processing (SAP) benchmark. Comprehension questions about the critical dependency in a subset of trials.\vspace{-1ex}
\item \citet{timkey_huang_oh_prasad_arehalli_linzen_dillon_2025}: A large-scale eye-tracking study (n = 367) replicating \citet{huang2024large} with the exact same experimental materials.\vspace{-1ex}
\item \citet{huang2021causes}, Expt.~1: An eye-tracking study on NP/Z garden paths (n = 120). Comprehension questions about the critical dependency in each trial.\footnote{Both \citet{huang2021causes} and \citet{fujita2025maze} additionally manipulated the gender (mis)match between reflexive pronouns and their antecedents. The critical reflexive region appeared after the regions we analyze here, but sentences did already differ in earlier regions, affecting the surprisal values at our regions of interest.}\vspace{-1ex}
\item \citet{fujita2025maze}: A Maze experiment on NP/Z garden paths (n = 60). No end-of-sentence task.\footnote{The participants in this study were proficient non-native speakers as opposed to the native speakers recruited for the other studies.}\vspace{-1ex}
\item \citet{paape2022conscious}: Three experiments (1 $\times$ SPR, 2 $\times$ BSPR) on NP/Z and MV/RR garden paths (n = 97/96/100). Grammaticality judgments (\textit{accept/reject}) in each trial.
\end{itemize}

For the eye-tracking and BSPR data, our measure of interest is first-pass reading time, that is, the time registered from first entering the region of interest to first leaving it. The data from all seven experiments were preprocessed and combined into one data set in R \citep{R}. Trials with reading times smaller than 150\,ms or greater than 5000\,ms at either the pre-critical, critical or spillover region were removed. This affected 2\% of the data. All models were fitted in Stan \citep{stan} via the CmdStanR interface \citep{cmdstanr}. Weakly informative priors based on previous work were used \citep{nicenboim2025introduction}. For each model, two chains with 2000 iterations each were run, with the first 1000 iterations being discarded as burn-in. The pathfinder algorithm \citep{zhang2022pathfinder} was used to choose starting values for the MCMC chains. $\hat{R}$ values were monitored for possible indications of non-convergence \citep{rhat}.\footnote{Some models initially showed evidence of bimodal posteriors, with the $shift$, $\sigma_2$ and global reading time intercept parameters falling in different basins across chains. As the theoretically more interesting parameters, such as garden-path probability and reanalysis cost, showed similar values across modes, we slightly modified priors so that all chains converged to the same mode.} Claude Sonnet 5 (Anthropic PBC) was used to audit R and Stan program code.

For the surprisal-based regression model, we computed surprisal values from three language models available via HuggingFace \citep{wolf2019huggingface}: the 124M variant of GPT-2 \citep{radford2019language}, the 2.8B variant of Mamba \citep{gu2023mamba} and the 4.3B variant of \textsc{rwkv} \citep{peng2023rwkv}. The Mamba and \textsc{rwkv} language models are recurrence-based, and have been found to have better fit to human data across different experimental paradigms, including SPR and Maze \citep{michaelov2024revengefallenrecurrentmodels}. Suprisal values were computed for the pre-critical, critical and spillover regions of each sentence for each language model. 

The surprisal slope was allowed to vary hierarchically between participants. For both the MPT and surprisal models, we also entered region length in characters as a predictor\footnote{Another ``nuisance'' predictor that is often included in the literature is unigram frequency (e.g., \citealp{van2021single,huang2024large}). However, given that disentangling surprisal and frequency effects is not our concern here, we ignore frequency and let the random effects absorb the associated variance.}, as the regions of interest had very different lengths across studies. Regularizing LKJ priors \citep{lewandowski2009generating} with $\eta = 4$ were used for the correlation matrices of the random effects, which downweight extreme correlations.

Apart from the base MPT and surprisal models, we also implemented a number of additional models, so that seven model types were compared overall:

\begin{enumerate}
\item \textsc{Baseline}: A multivariate regression model with categorical predictors (condition, construction), plus region length and random effects.\vspace{-1ex}
\item \textsc{MPT}: The MPT model as described above.\vspace{-1ex}
\item \textsc{Surprisal}: A linear regression model with GPT-2/Mamba/\textsc{rwkv} surprisal as a predictor, plus region length and random effects.\vspace{-1ex}
\item \textsc{MPT+surprisal}: The MPT model as described above, plus a linear surprisal effect on reading times for each region.\vspace{-1ex}
\item \textsc{MPT+surprisal-as-probability}: The \textsc{MPT+surprisal} model, with the categorical condition slope on $p\_gp$ replaced with the surprisal of the critical disambiguating region.\vspace{-1ex}
\item \textsc{MPT+surprisal-as-cost}: The \textsc{MPT+surprisal} model, with an added slope for the surprisal of the critical region on $gp\_cost$.\vspace{-1ex}
\item \textsc{MPT+surprisal-all}: The \textsc{MPT+surprisal} model, with all categorical predictors inside the MPT replaced with the surprisal of the critical region.
\end{enumerate}

All models with surprisal slopes were fitted with GPT-2, Mamba, and \textsc{rwkv} surprisal, bringing the total number of compared models to 17. All models are multivariate; the dependent variables are reading times, regression probabilities, and grammaticality judgments/question responses. A shifted lognormal likelihood is used for the reading time distributions. For the surprisal and baseline models, regression probabilities and grammaticality judgments/question responses are treated as Bernoulli variables, and separate surprisal slopes are estimated for each dependent variable (reading times, regression probabilities, judgments, question responses). 

The models do not directly receive information about which study a given data point comes from. Participants and sentences are assumed to be sampled from Gaussian random-effects distributions that are shared across all experiments. However, we do hard-code the experimental method, which affects the amount of shift (non-decision time). We also hard-code whether regressions were possible in the current trial, and let baseline regression probability and regression cost vary between paradigms. The probability of postponed reanalysis and the probability of overt reanalysis are also allowed to vary between paradigms. We hard-code the end-of-trial task (grammaticality judgments vs.~comprehension questions) to determine whether pragmatic inferences can occur or not. Finally, within MV/RR trials, we hard-code the disambiguation type (semantic vs.~syntactic), as explained with regard to (\ref{x2}) above. 

\subsection*{Model comparisons}

We used the loo R package \citep{loo1} to carry out model comparisons via cross-validation. The aim of cross-validation is to quantify how well a given model predicts previously unseen data points. To do this, a subset of the data is held out, the model is fitted to the remaining data, and the expected log pointwise predictive density ($\widehat{elpd}$) of the held-out data is computed. This procedure is repeated for different unique subsets of the data, until the entire data set is covered. Models with higher $\widehat{elpd}$ have better predictive (out-of-sample) fit. Quantifying and comparing out-of-sample fit rather than in-sample fit has the advantage of penalizing overfitting: models that are too tightly optimized to the fitted data will usually not predict unseen data well. Given that leave-one-out cross-validation is computationally expensive, the loo package computes an approximation via importance sampling \citep{loo2}. In general, moving away from optimizing in-sample fit towards optimizing out-of-sample fit is an important goal for cognitive science, as models with good out-of-sample fit should provide better generalizability \citep{yarkoni2022generalizability}.

\begin{figure}[h]\centering
 \includegraphics[scale=.45]{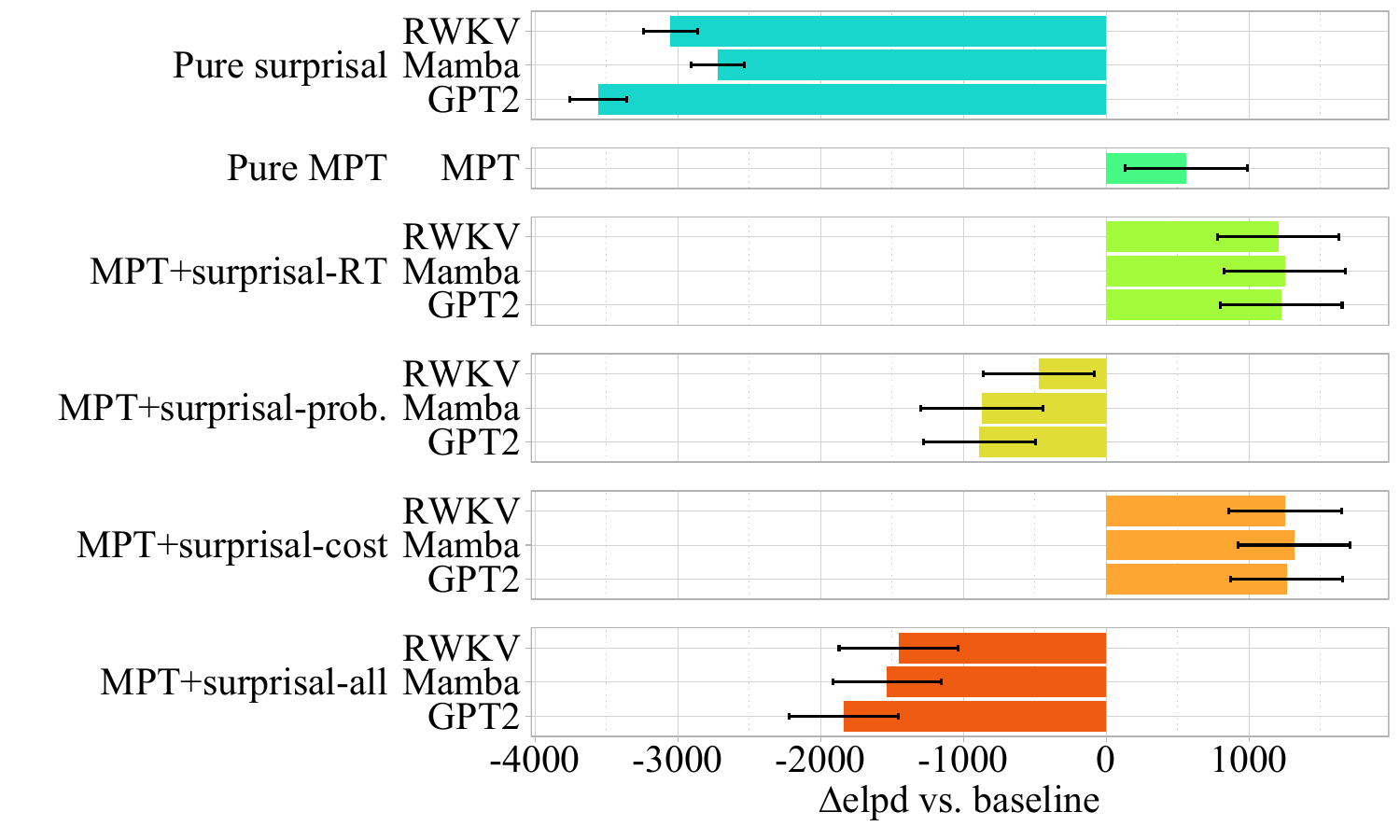}
 \caption{Differences in $\widehat{elpd}$ between each model and the \textsc{baseline} model (multivariate with categorical predictors). Higher $\Delta\widehat{elpd}$ means better predictive performance compared to the baseline. Error bars represent 95\% confidence intervals.\label{diffs}}
\end{figure}

Our cross-validation procedure compared each non-baseline model to the baseline. Figure \ref{diffs} shows the results. The main observations are:

\begin{enumerate}
\item Purely surprisal-based models have worse predictive fit than the baseline model with categorical predictors.\vspace{-1ex}
\item The base MPT model outperforms the baseline categorical model.\vspace{-1ex}
\item Adding surprisal of the disambiguating region to the MPT model as a predictor of RT and/or garden-path cost further improves predictive fit.\vspace{-1ex}
\item Adding surprisal of the disambiguating region as a predictor of garden-path probability or replacing all categorical predictors with surprisal reduces predictive fit compared to the base MPT model.
\end{enumerate}

Because the pure surprisal model is the main theoretical competitor to the MPT model that we are interested in, we will zoom in on where in the data the MPT model gets its main predictive advantage from. This is a useful approach when doing model comparisons, as the theoretical assumptions of each model should be reflected in how well it is able to predict data that fit the assumptions (e.g., \citealp{nicenboim2018models}). For instance, it might be that the advantage of the MPT model is mainly driven by trials with very long reading times, which would reflect the assumption that reanalysis is a categorically distinct, costly process. Furthermore, even though the MPT model outperforms the pure surprisal models on average, it might be that there are pockets of data for which the surprisal models have a predictive advantage.

\begin{figure}[h]\centering
\begin{subfigure}[t]{.55\textwidth}
 \includegraphics[width=\textwidth]{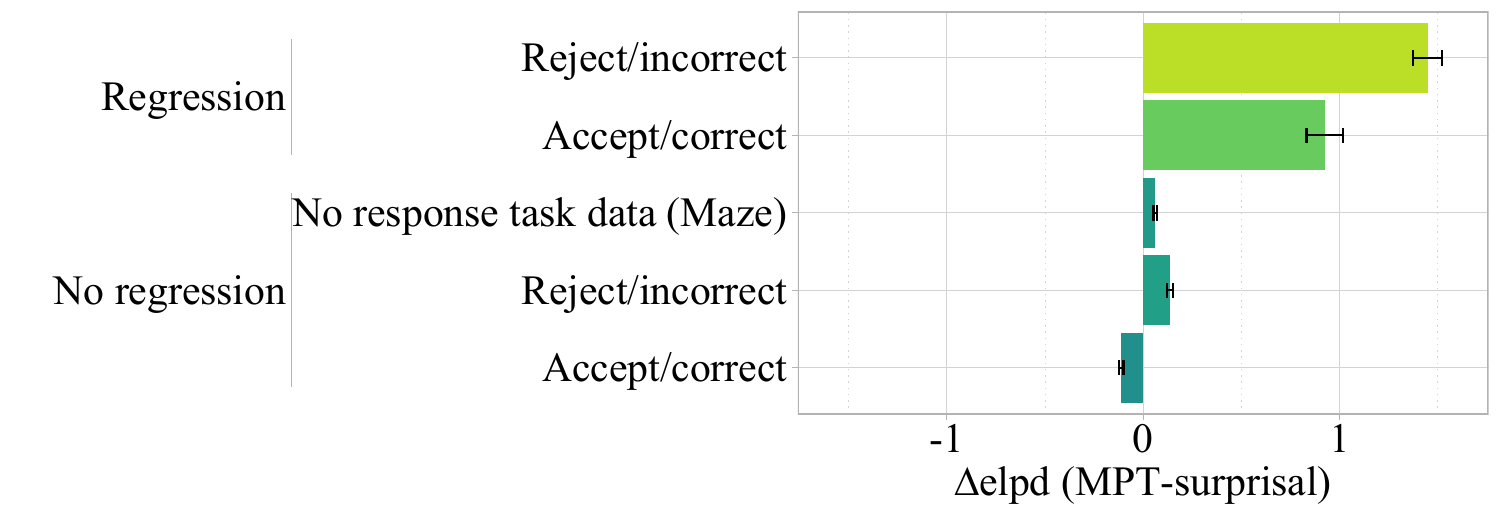}
 \caption{By trial type}
\end{subfigure}\hfill
\begin{subfigure}[t]{.4\textwidth}
 \includegraphics[width=\textwidth]{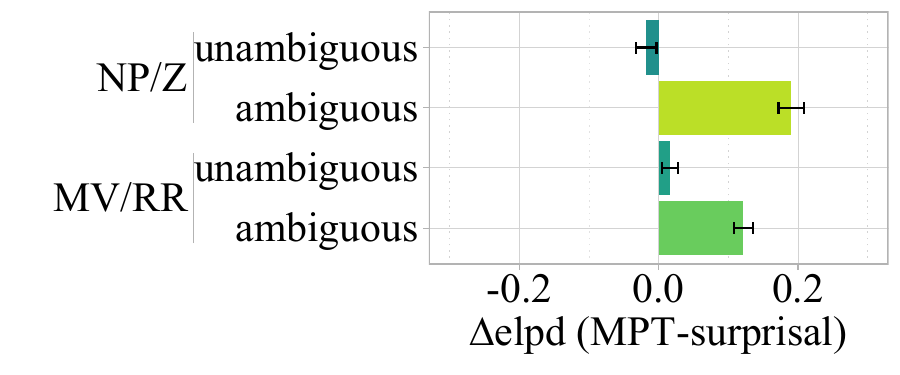}
  \caption{By construction and condition}
\end{subfigure}
\par\medskip %
\begin{subfigure}[t]{.4\textwidth}\centering
 \includegraphics[width=\textwidth]{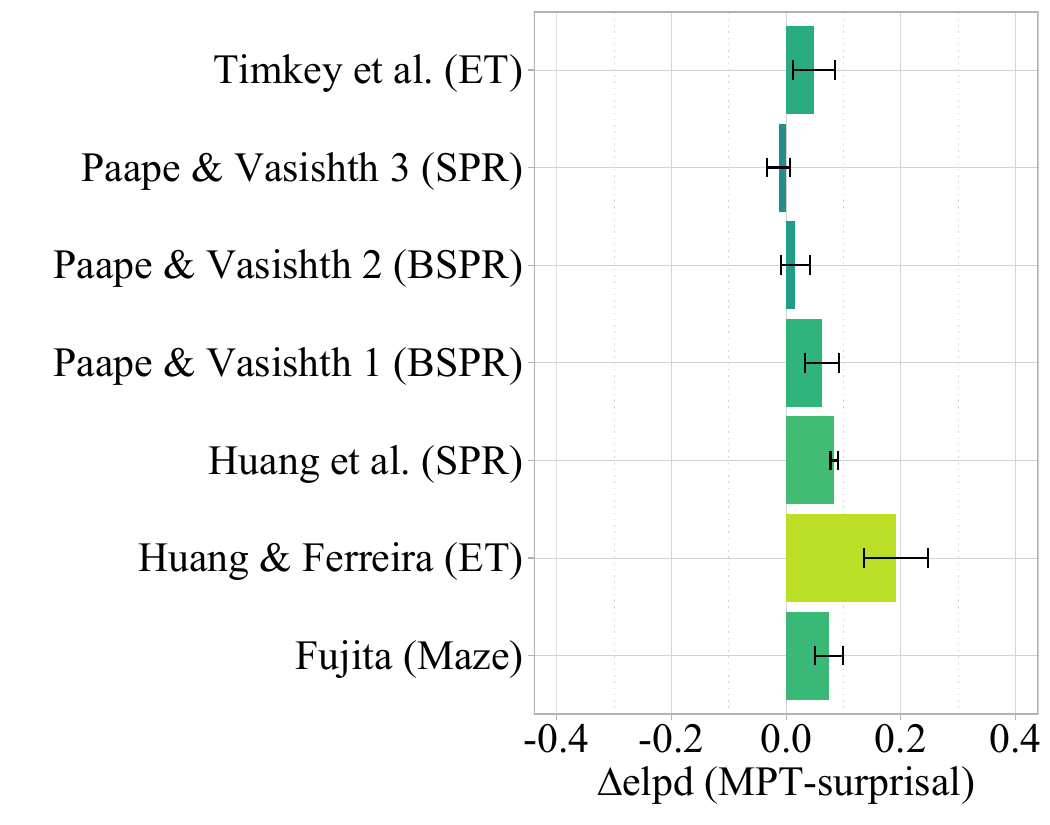}
  \caption{By experiment}
\end{subfigure}\hfill
\begin{subfigure}[t]{.55\textwidth}\centering
 \includegraphics[width=\textwidth]{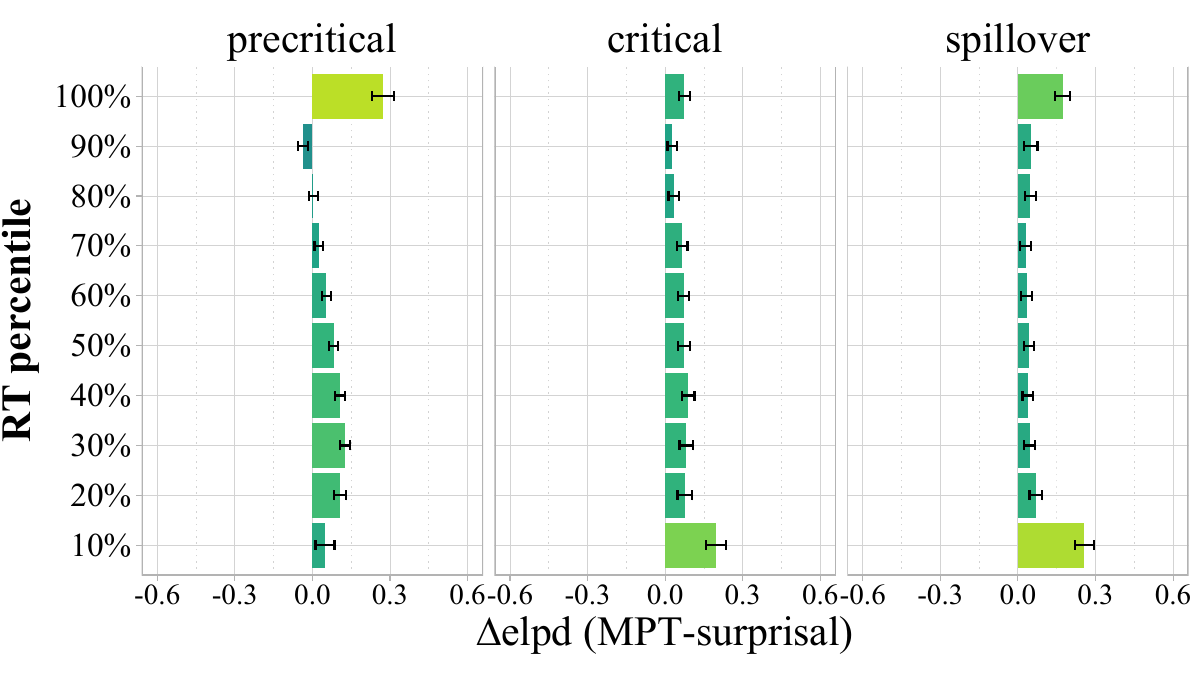}
  \caption{By region and RT percentile}
\end{subfigure}
\caption{Detailed breakdown of differences in $\widehat{elpd}$ between the Surprisal-Mamba and MPT+Mamba-cost models. The more positive the value, the more MPT+Mamba-cost outperforms Surprisal-Mamba.\label{comps}}
\end{figure}

Figure \ref{comps} shows the difference in $\widehat{elpd}$ between the MPT+Mamba model and the best-performing pure surprisal model  across a variety of variables. The overall pattern is that the MPT advantage is largest when processing difficulty is high: the MPT model has better predictive fit for trials with regressions and/or rejections (trials with either negative grammaticality judgments or incorrect question responses), and for the ambiguous compared to the unambiguous conditions. The MPT model also outperforms surprisal across all reading-time percentiles, but especially strongly for the top decile. Surprisal outperforms the MPT model for ``accept'' trials without regressions, for unambiguous MV/RR sentences, and for the 10\% shortest reading times in the precritical region. Surprisal also slightly outperforms the MPT model for the \citet{timkey_huang_oh_prasad_arehalli_linzen_dillon_2025} eye-tracking data, which may appear surprising given the authors' suggestion that surprisal may not or may only partly capture structural processing. However, \citeauthor{timkey_huang_oh_prasad_arehalli_linzen_dillon_2025}'s conclusions were based mainly on measures of rereading, which, apart from regression probabilities, were not included in our model.

The finding that very long reading times are a strong driver of the MPT advantage is consistent with a large reanalysis cost being the most distinctive feature of garden-path sentences (e.g., \citealp{largerare}). However, it appears somewhat surprising that long reading times in the precritical region are also better explained by the MPT model, given that no reanalysis is expected to occur in this region. The relevant insight here is that a single $\widehat{elpd}$ value is always computed for the entire trial: if reading times for the pre-critical region are correlated with other measures within the trial, seeing high $\widehat{elpd}$ for a given decile does not entail a causal relationship. Indeed, a look at the data reveals that high pre-critical RTs are correlated with a higher incidence of regressions at the critical and post-critical regions, as well as with more rejections. The MPT model can account for these correlations via the in(attention) mechanism, where attentive reading leads to longer, more variable reading times as well as more regressions and rejections.

The role of ``reject'' and regression trials in the MPT advantage, as well as the comparatively stronger impact of the NP/Z ambiguity, are in line with the assumption that some structures are categorically more difficult to reanalyze than others (e.g., \citealp{sturt1999structural}), as well as with the assumption that the parsing of garden-path sentences can require rereading and/or sometimes fail outright. Surprisal, being essentially a theory of successful parsing, has difficulty capturing these cases \citep{timkey_huang_oh_prasad_arehalli_linzen_dillon_2025}.

\subsection*{Parameter estimates and predictive fit of the MPT+Mamba-cost model}

Table \ref{t1} shows the parameter estimates for the MPT+Mamba-cost model, the model with the overall best predictive performance. This model combines the MPT architecture with an additional linear surprisal effect on reading times across all three regions (pre-critical, critical, spillover), as well as an additional effect of surprisal on garden-path cost. Graphical posterior predictive checks for the MPT+Mamba-cost are given in the appendix.

\begin{table}[htb]\centering
\caption{Selected parameter estimates (back-transformed upper and lower bounds of 95\% CrI) for the MPT model. Non-decision time ($shift$) is saturated: it is a fraction of each participant's fastest RT, and the estimate sits at the ceiling. The same holds for $\beta_{10}$, the difference between the Maze and (B)SPR ceiling values. By contrast, $p\_attentive$ is very tightly determined by the data. For the rows marked $\dagger$, we report the range covering 95\% of by-participant random effects instead of the CrI.\label{t1}}
\vskip 0.12in
\def\arraystretch{1.1}%
\begin{tabular}{lrrl}
\hline
\multicolumn{4}{c}{Costs (intercepts)}\\
\hline
parameter&low&high&interpretation\\
$shift^\dagger$&148\,ms&433\,ms&Non-decision time (SPR/BSPR)\\
$att\_cost$&232\,ms&255\,ms&Cost of attention\\
$gp\_cost$&6\,ms&10\,ms&Cost of garden-pathing\\
$reanalysis\_cost$&202\,ms&227\,ms&Covert reanalysis cost\\
$regression\_cost$&229\,ms&294\,ms&Regression cost (BSPR)\\
\hline
\multicolumn{4}{c}{Probabilities (intercepts)}\\
\hline
parameter&low&high&interpretation\\
$p\_attentive^\dagger$&0.88&0.99&Prob.~of attentive reading\\
$p\_gp$&0.31&0.33&Prob.~of garden-pathing (unamb.~conditions)\\
$p\_overt$&0.03&0.07&Prob.~of overt reanalysis (BSPR)\\
$p\_postpone$&0.42&0.46&Prob.~of postponing reanalysis (ET)\\
$p\_success\_o$&0.63&0.70&Success prob.~of overt reanalysis\\
$p\_success\_c$&0.23&0.25&Success prob.~of covert reanalysis\\
$p\_infer$&0.25&0.43&Prob.~of pragmatic inference\\
$p\_base\_regress$&0.05&0.06&Baseline regression prob.~in BSPR\\
\hline
\multicolumn{4}{c}{Slopes}\\
\hline
parameter&low&high&interpretation\\
$\beta_1$&+0.48&+0.56&Ambiguity effect on GP prob.\\
$\beta_2$&$-$0.03&+0.12&Ambiguity effect MV/RR vs.~NP/Z\\
$\beta_3$&+82\,ms&+117\,ms&Covert reanalysis cost MV/RR vs.~NP/Z\\
$\beta_4$&+0.51&+0.62&Overt reanalysis success MV/RR vs.~NP/Z\\
$\beta_5$&+0.27&+0.31&Covert reanalysis success MV/RR vs.~NP/Z\\
$\beta_6$&+0.30&+0.32&Baseline regressions ET vs.~BSPR\\
$\beta_7$&+0.03&+0.10&Overt reanalysis ET vs.~BSPR\\
$\beta_8$&+0.03&+0.08&Postpone MV/RR vs.~NP/Z\\
$\beta_9$&+0.21&+0.24&Postpone (B)SPR vs.~ET\\
$\beta_{10}$&$-$0.26&$-$0.15&Postpone Maze vs.~ET\\
$\beta_{11}$&+224\,ms&+224\,ms&Non-decision time Maze vs.~(B)SPR\\
\end{tabular}
\end{table}

The estimates are in line with expectations based on the literature: a much higher probability of garden-pathing in the ambiguous conditions compared to the unambiguous conditions, a comparatively large cost of covert reanalysis, and a higher probability of successful reanalysis in MV/RR compared to NP/Z sentences. Interestingly, garden-pathing is less likely but covert reanalysis is both more costly and more successful in MV/RR sentences, possibly suggesting that incorrect structural commitments that occur less often take more time to recover from but are also less failure-prone. 

Compared to the estimates of \citet{paape2026deconstructing}, the estimated probability of overt reanalysis is much lower. This is likely due to the inclusion of the precritical region, which leads to a cleaner estimate of the baseline regression probability. \citeauthor{paape2026deconstructing} modeled only the critical and spillover regions, where it is difficult to disentangle reanalysis-driven regressions from regressions due to other factors such as motor-error correction \citep{mitch}. In terms of differences between reading paradigms, eye tracking is estimated to have a much higher baseline regression probability than BSPR. This is in line with the observation that eye tracking tends to often show short, corrective regressions even in unambiguous sentences, while regressions in BSPR are more deliberate and occur less often overall \citep{paape2022conscious}. 

Regarding processing spillover, which is modeled via postponed reanalysis, eye tracking is positioned squarely between BSPR and Maze, which has the smallest amount of spillover. Again, this pattern is in line with claims in the literature \citep{boyce2023maze}. Non-decision time is much longer in Maze compared to BSPR, consistent with the assumption that having to eliminate the non-appropriate continuation in each region takes more time than simply making the decision to move forward (or backward) through the text.

\section*{General discussion}

The current work had two main aims. The first was to apply an extension of the \citet{paape2022estimating} multinomial processing tree (MPT) model of garden-pathing to data from different reading paradigms, namely eye tracking, uni- and bidirectional self-paced reading, and Maze. The MPT model makes a crucial implicit assumption in psycholinguistics explicit: latent cognitive processes such as attention and syntactic reanalysis are underlyingly shared between reading paradigms but are mediated by paradigm-specific factors such as deliberate (SPR, BSPR, Maze) versus largely automatic (eye tracking) control over the reading process or by the ability or inability to reread sentences. We have shown that the MPT model outperforms a baseline multivariate model across paradigms in terms of predictive fit, and that the parameter estimates are in line with claims in the literature, such as Maze being less prone to spillover than SPR.

The second aim was to contribute to the ongoing discussion about the role of surprisal in garden-path processing, and possibly in sentence processing more generally (e.g., \citealp{staub2025predictability,slaats2025}). In their recent eye-tracking study on garden paths, \citet{timkey_huang_oh_prasad_arehalli_linzen_dillon_2025} have shown that surprisal does an adequate job at predicting forward reading time in the absence of a regression, but fails to predict rereading-related behavior. They concluded that surprisal was likely capturing word recognition but not necessarily word integration in context, or only those integrations that were immediately successful. Previous work using self-paced reading data has also shown that surprisal fails to capture the full magnitude of reading slowdowns in garden path sentences \citep{pvc,van2021single,huang2024large}. However, \citet{yoshida-etal-2026-existence} have recently shown that fine-tuning GPT-2 on the \citet{huang2024large} experimental materials results in improved fit to the human data.

Our current work adds further nuance to the debate by showing that both a smooth surprisal function \textit{and} the assumption of qualitatively distinct trial types --- attentive vs.~inattentive processing, reanalysis vs.~no reanalysis --- are needed to obtain an optimal predictive fit to human data. While the base MPT model clearly outperforms pure surprisal models, a further boost in predictive performance was obtained by including surprisal as an additional predictor of reading time and initial garden-path cost. We interpret this finding as being in line with \citeauthor{timkey_huang_oh_prasad_arehalli_linzen_dillon_2025}'s conclusions: surprisal may either only capture word recognition while the MPT part of the model captures word integration, or surprisal may capture successful integration while the MPT captures specialized recovery mechanisms. Letting surprisal affect other parts of the MPT cascade, such as the probability of garden-pathing or the garden-path cost, either failed to improve predictive fit or even decreased it, suggesting that surprisal may indeed not interact with these processes.

A crucial point is that we gave pure surprisal a generous chance at predicting ambiguity-induced slowdowns by fitting our models only to garden-path sentences. This was also the approach taken by \citet{pvc}. By contrast, other authors \citep{timkey_huang_oh_prasad_arehalli_linzen_dillon_2025,van2021single,huang2024large,mainakilaas2026revisiting} were more restrictive: they fitted their models only to filler sentences and then used the resulting surprisal slope to predict garden-path reading times. The only factor making our approach somewhat more restrictive than that of \citet{pvc} is our inclusion of the precritical region, which was intended to provide a baseline of ``normal'' reading times and regression probabilities. 

Another factor that presumably worked in favor of surprisal is that we only included first-pass reading times for BSPR and eye tracking, excluding rereading times. In the \citet{timkey_huang_oh_prasad_arehalli_linzen_dillon_2025} data, surprisal provided a substantially worse fit to measures that included rereading, suggesting that including them in our model would have reduced predictive fit as well. That being said, our point of comparison is the MPT model, whose performance on non-garden-path sentences and rereading measures has never been evaluated, so that it is difficult to predict which model would end up predicting the human rereading data better. We leave this question to future research.

Against the background of our results, what does the mismatch between LLM-derived surprisal and human processing difficulty in garden-path sentences tell us about how humans and language models process sentences? One plausible reason for the mismatch is that garden-path sentences \textit{are} categorically different from non-garden-path sentences, and that LLMs lack a dedicated mechanism that would parallel the slow, laborious, and sometimes conscious process of syntactic reanalysis seen in humans \citep{van2021single,paape2022estimating,pvc}. Another possible reason is that LLMs can simply maintain more concurrent parses than humans, and switch between them more easily upon disambiguation. However, \citet{timkey2026language} have recently demonstrated that reducing the number of parallel parses in an LLM increases the magnitude of garden-path effects (see also \citealp{mainakilaas2026}) but still does not result in human-like processing difficulty. 

The existence of specialized recovery mechanisms in humans would be in line with the more modular view of psycholinguistic processing that was widely taken in earlier work on garden-pathing (e.g., \citealp{marcus1980theory,pritchett1992grammatical,frazier1979comprehending}). If this conclusion holds, it would underscore recent concerns that LLMs have, on the one hand, become ``too superhuman'' to serve as accurate models of human language processing \citep{oh2023does,oh2025model}, but at the same time lack fundamental cognitive processes that humans possess \citep{lin2025six,schröder2025largelanguagemodelssimulate}. 

Regarding syntactic reanalysis specifically, some headway has recently been made by extracting information from internal LLM states and using them to derive sentence representations \citep{wang-etal-2025-extracting,li2024incremental}. The resulting models show improved fit to human garden-path processing data, suggesting that internal LLM states do capture some aspects of the specific difficulty that humans experience with these kinds of structures. This line of research promises to yield further insights into parallels between humans and LLMs in garden-path processing, but it does represent a marked shift away from the ``causal bottleneck'' assumption with regard to surprisal.

\section*{Conclusion}

Using data from four different reading paradigms --- eye tracking, uni- and bidirectional self-paced reading, and Maze --- we have shown that a single computational model can capture the processing of garden-path sentences by assuming a cascade of latent cognitive processes that are shared across paradigms. The processes are assumed to trigger nondeterministically, and cover (in)attention, adoption of an initial analysis, and either covert or overt reanalysis as components of reading behavior. Additional processes, namely guessing, failed reanalysis or pragmatic inference, explain misinterpretations or negative grammaticality judgments. We compared the predictive fit of the latent-process model to that of a simpler model based only on LLM-derived surprisal, and found that the latent-process model decisively outperformed surprisal. However, including surprisal as an additional component in the latent-process model improved fit over the base version. Overall, we have demonstrated that multi-paradigm data is a useful resource for cognitive modeling, and that surprisal is necessary but not sufficient to capture English garden-path effects, which are best modeled by assuming a mixture of qualitative different trials involving different processes.

\printbibliography

\section*{Appendix: Posterior predictive checks for the MPT+Mamba-cost model}

\begin{figure}[htb]
\centering
 \includegraphics[scale=.7]{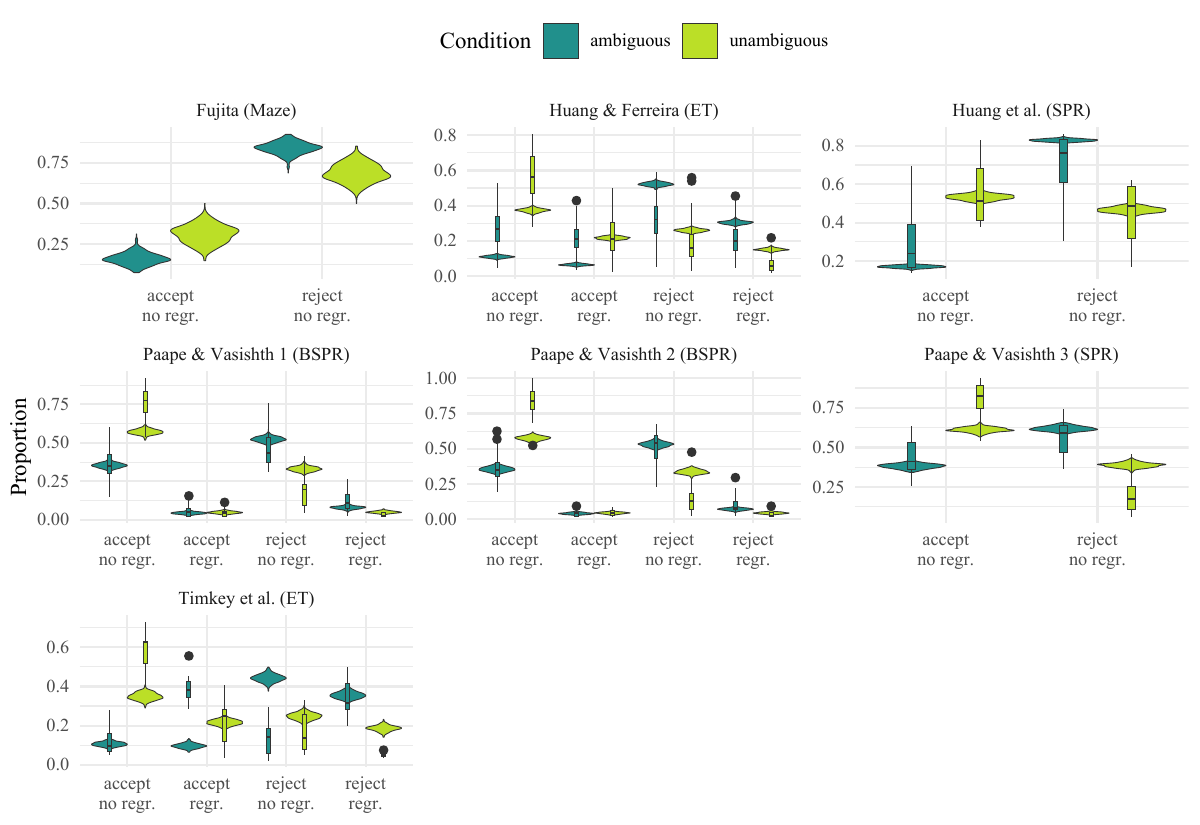}
 \caption{Observed proportions of trial types (boxplots) and predicted proportions from the MPT+Mamba-cost (violins) for NP/Z sentences. Correct question responses are grouped as \textit{accept}, incorrect question responses are grouped as \textit{reject}. Circles show boxplot outliers in the observed data.\label{ppc1}}
\end{figure}

\begin{figure}[htb]
\centering
 \includegraphics[scale=.7]{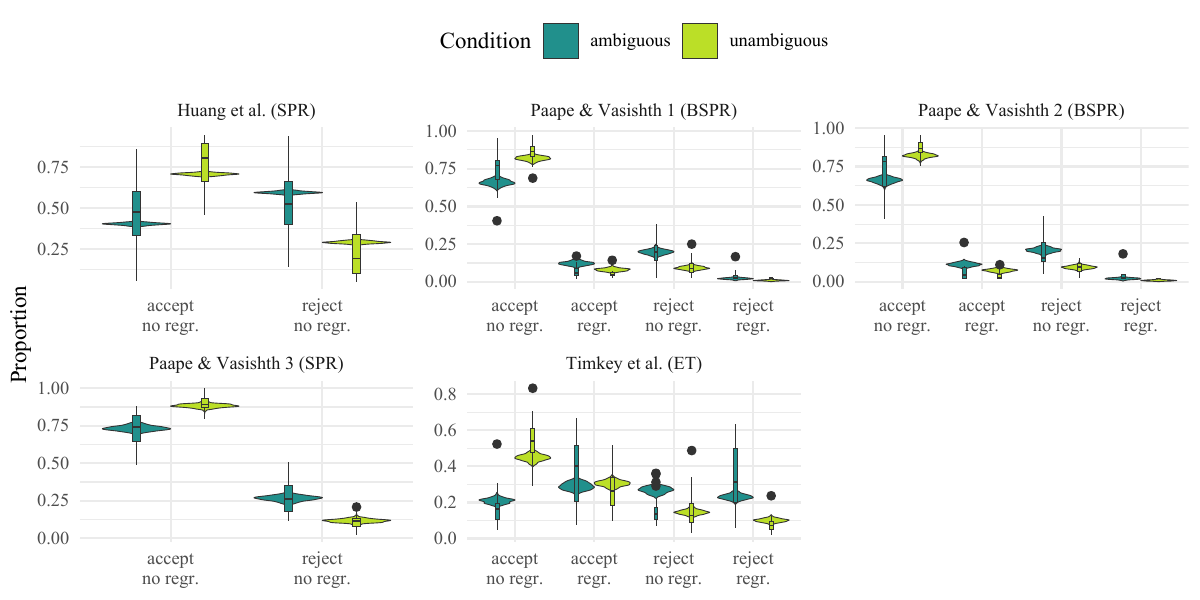}
 \caption{Observed proportions of trial types (boxplots) and predicted proportions from the MPT+Mamba-cost model (violins) for MV/RR sentences. Correct question responses are grouped as \textit{accept}, incorrect question responses are grouped as \textit{reject}. Circles show boxplot outliers in the observed data.\label{ppc2}}
\end{figure}

Graphical posterior predictive checks for trial-type proportions are shown separately for NP/Z sentences (Figure \ref{ppc1}) and MV/RR sentences (Figure \ref{ppc2}). The posterior predictive distribution is obtained by generating new data from the fitted model with each iteration of the sampling algorithm. Posterior predictive checks then compare the predicted data to the actual data and show whether the model can reproduce the empirical patterns (e.g., \citealp{gelman1996posterior}). Recall that there is no trial-type data for the \citet{fujita2025maze} Maze experiment, as there was no end-of-trial task and no rereading.

Overall, the MPT+Mamba-cost model is able to capture the qualitative patterns in the data. The one exception is a pattern in the \citet{timkey_huang_oh_prasad_arehalli_linzen_dillon_2025} data for NP/Z sentences, where ambiguous sentences are more likely to have correct question responses (``accept'') after a regression, and are \textit{less} likely to have incorrect question responses in the \textit{absence} of a regression than the model predicts. The mismatch between the observed and predicted data for this specific experiment and sentence type is likely driven by patterns that dominate in the rest of the data set, and suggests that the MPT+Mamba-cost model is not flexible enough to account for the deviation. This being said, too much flexibility in the sense that any arbitrary pattern can be predicted is not desirable in a cognitive model \citep{Pitt2002WhenAG}, and mismatches between model predictions and data can inform theory building. In the current case, further research is needed into the interaction between rereading and comprehension accuracy in NP/Z  compared to MV/RR sentences, whose patterns the model is able to accurately capture.

\end{document}